\documentclass{article}

\usepackage{PRIMEarxiv}
\usepackage{authblk}
\usepackage{amssymb}        
\usepackage[utf8]{inputenc} 
\usepackage[T1]{fontenc}    
\usepackage{hyperref}       
\usepackage{url}            
\usepackage{booktabs}       
\usepackage{amsfonts}       
\usepackage{nicefrac}       
\usepackage{microtype}      
\usepackage{lipsum}
\usepackage{subcaption}
\usepackage{amsmath}
\usepackage{dsfont}         
\usepackage{fancyhdr}       
\usepackage{graphicx}       
\usepackage{svg}
\usepackage{algorithm}
\usepackage{algorithmic}
\usepackage{cleveref}
\usepackage[utf8]{inputenc}
\graphicspath{{media/}}     

\title{A Physics Informed Neural Network (PINN) Methodology for Coupled Moving Boundary PDEs}
\author[a]{Shivprasad Kathane}
\author[a,b,1]{Shyamprasad Karagadde}
\affil[a]{Centre for Machine Intelligence and Data Science, Indian Institute of Technology Bombay, Mumbai 400076, India}
\affil[b]{Department of Mechanical Engineering, Indian Institute of Technology Bombay, Mumbai 400076, India}
\affil[1]{Corresponding author, email: s.karagadde@iitb.ac.in}

\begin{document}

\maketitle

\begin{abstract}
Physics-Informed Neural Network (PINN) is a novel multi-task learning framework useful for solving physical problems modeled using differential equations (DEs) by integrating the knowledge of physics and known constraints into the components of deep learning. A large class of physical problems in materials science and mechanics involve moving boundaries, where interface flux balance conditions are to be satisfied while solving DEs. Examples of such systems include free surface flows, shock propagation, solidification of pure and alloy systems etc. While recent research works have explored applicability of PINNs for an uncoupled system (such as solidification of pure system), the present work reports a PINN-based approach to solve coupled systems involving multiple governing parameters (energy and species, along with multiple interface balance equations). This methodology employs an architecture consisting of a separate network for each variable with a separate treatment of each phase, a training strategy which alternates between temporal learning and adaptive loss weighting, and a scheme which progressively reduces the optimisation space. While solving the benchmark problem of binary alloy solidification, it is distinctly successful at capturing the complex composition profile, which has a characteristic discontinuity at the interface and the resulting predictions align well with the analytical solutions. The procedure can be generalised for solving other transient multiphysics problems especially in the low-data regime and in cases where measurements can reveal new physics.
\end{abstract}


\section{Introduction}\label{sec:intro}
\paragraph{}Modelling and forecasting the dynamics of complex multiphysics systems is a crucial challenge with relevance to several engineering applications. Traditionally, for simulating such systems, standard numerical methods are used to solve the partial differential equations (PDEs) that model the underlying physics of the problem. However, these approaches tend to be based on time-marching schemes to obtain solution at a later time, and find it difficult to solve certain PDEs that represent significant non-linearity, convection dominance, or shocks and especially if the system consists of multiple coupled PDEs\cite{SciML}. Further, several emerging approaches are based on machine learning, especially involving deep neural networks, \cite{papreview, hodreview}. But conventional deep learning is data-intensive and ignores any available knowledge of conditions or process information. The Physics-Informed Neural Network (PINN), first introduced in \cite{PINN}, is a novel multi-task learning artificial intelligence (AI) framework with the dual objective of fitting the observed data and minimising the residual on the PDEs and constraints. Such a hybrid approach ensures that the trained model can recognise patterns from the data (observational bias), respect the conservation laws (inductive bias) and can be used to predict the dynamics of the system (learning bias) which forms the basis of the field of physics-informed machine learning \cite{PIML}. PINNs are able to approximate PDE solutions by training a neural network to minimise a loss function which includes the terms reflecting the initial conditions (at start time) and boundary conditions (in space), and the PDE residual at selected points in the domain (collocation data) which acts as a penalising term restricting the space of acceptable solutions. Also, the training of a PINN is an unsupervised mesh-free strategy that does not require results from prior simulations or experiments (necessary for a general supervised deep learning problem) and is free from step limitations (that otherwise arise in conventional marching-based numerical simulations) and once trained, it can be readily used for predictions. Further, it also presents a systematic possibility of utilizing sporadic measurements as a part of the solution methodology. These merits make it promising to utilise the framework for solving complex problems.

\paragraph{} The PINN introduced in \cite{PINN}, in its native format, can be applied to address different challenges posed by various problems, listed as follows. PINN has been used to solve non-linear problems such as the Burger's equation where it was well able to capture the shock formation \cite{Burgers}. The framework has also used for a non-linear multiphysics system with complex geometries such as for solving the coupled diffusivity and Biot's equations \cite{Biot}. Authors in \cite{Stokes} outline strategies for solving problems having small parameter values and interface discontinuity issues as they solve the coupled Stokes-Darcy Equation using PINN. The flexibility of the framework has been explored for real-world scenarios in the domain of heat transfer, for example to solve the two-phase Stefan problem with missing thermal boundary conditions \cite{Stefan} which involved temperature prediction while tracking the interface movement. PINN has also been employed for modelling manufacturing processes such as the thermochemical curing process of a composite-tool system \cite{Curing} and the solidification of aluminium in a graphite mold \cite{MLAM}. While the former addressed the vastly different behaviours of conductive heat transfer and resin cure kinetics, the latter resolved the heat transfer coupled with the thermodynamics of phase transition. A problem of considerably larger complexity is the binary alloy solidification \cite{book1}, primarily due to the fact that a discontinuity in the composition exists at the solidifying interface, as opposed to a continuous thermal distribution. This is an example of a transient multi-component multi-phase multi-physics system. As the alloy solidifies with time, it is required to predict the temperature and composition distribution in the two phases along with the interface position. The physics is governed by four PDEs along with two additional constraints at the moving interface, and the corresponding initial and boundary conditions. The interface introduces discontinuities in the composition field directly, and in the gradient of the temperature field. These pose significant challenges from the perspective of modelling using PINN. A single standard PINN would find it difficult to optimise and converge to a solution in presence of about 15 loss terms while predicting three different variables across two phases. Thus, a strategy with a modified PINN-based methodology is needed to solve such a problem.

\paragraph{}In this paper, the next section \ref{sec:problem} describes the problem in detail listing the governing partial differential equations and constraints valid for the specified problem geometry. Section \ref{sec:solution} explains the solution methodology in detail, first, using the standard PINN approach with the associated loss terms. The following sub-section discusses advances in PINN approaches which is followed by a sub-section describing the final implemented methodology. Section \ref{sec:results} consists of results and figures for the final procedure implemented which show predictions that align well with the analytical solution. The significance of the results and the highlights of the methodology are discussed in section \ref{sec:conc}. To summarise the main contribution of this paper, we study a benchmark problem consisting of multiple complexities and investigate advances in PINN approaches to come up with a comprehensive approach by uniting strategies to accelerate the learning process. We use separate networks for each variable, treat each phase separately, and propose the alternate application of causal training and adaptive weighting of loss functions, and train in a sequentially-focused manner.

\section{Problem Description}\label{sec:problem}
\begin{figure}
  \centering
  \begin{subfigure}{0.64\textwidth}
    \includegraphics[width=\linewidth]{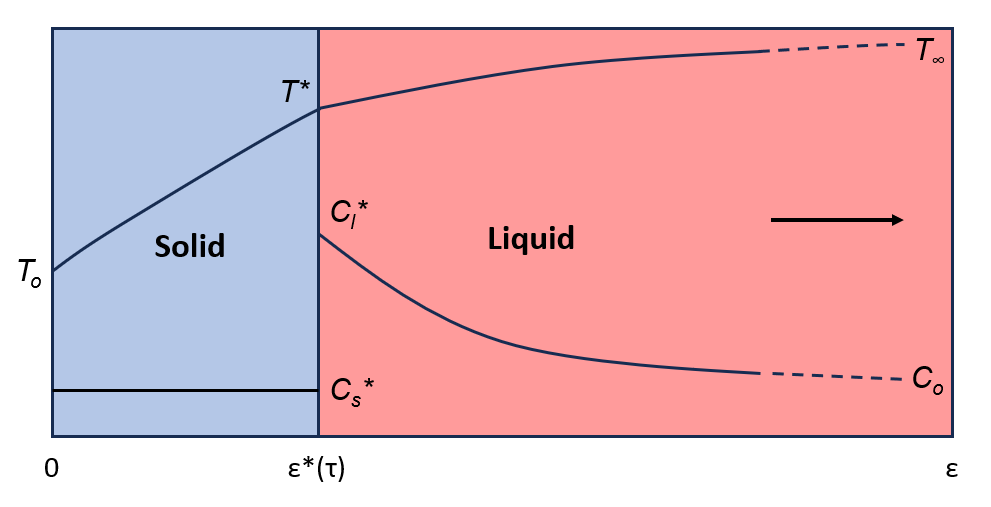}
    \label{fig:subfig1a}
  \end{subfigure}
  \hspace*{\fill}
  \begin{subfigure}{0.34\textwidth}
    \includegraphics[width=\linewidth]{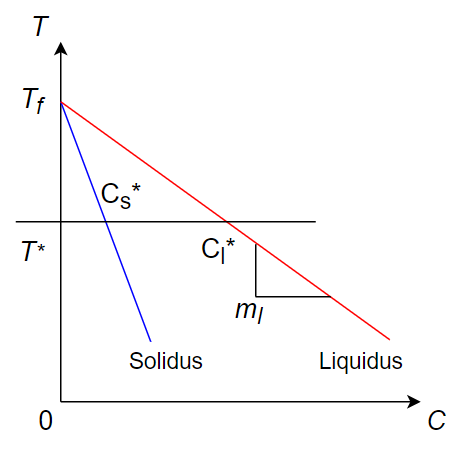}
    \label{fig:subfig1b}
  \end{subfigure}
  \caption{(a) Indicative thermal and compositional profiles in a one-dimensional system undergoing solidification of a superheated alloy melt. The left side boundary is subjected to cooling. The two phases and the sharp boundary separating them are shown at an intermediate stage. (b) the associated phase diagram showing equilibrium phase transition lines}
  \label{fig:fig1}
\end{figure}

\paragraph{}The binary alloy solidification problem consists of determining the temperature $T$ and composition $C$ distribution in the two phases which are separated by a moving interface. This implies solving the system of equations for the heat conduction problem and the corresponding solute problem while tracking the interface movement. The equations of thermal and species diffusion in the solid and liquid phases are written separately with the correspending interface balance conditions (Dantzig and Rappaz \cite{book1}), as represented schematically in Fig \ref{fig:fig1}. The left portion of the alloy is solid, while the right portion is liquid. The solid-liquid interface is moving towards right as solidification proceeds. It is assumed that all the thermal properties are constant and equal for the two phases and there is no fluid flow or internal heat generation. We present the dimensionless system of PDEs below. The reader is referred to \cite{book1} for further details. 

The heat equations for the two phases are given by the following PDEs:
\begin{equation}\quad \frac{\partial\theta_s}{\partial\tau} = \frac{\partial^2\theta_s}{\partial\epsilon^2}\quad for\quad \epsilon \in [0,\epsilon^*]\end{equation}
\begin{equation}
\quad \frac{\partial\theta_l}{\partial\tau} = \frac{\partial^2\theta_l}{\partial\epsilon^2}\quad for\quad \epsilon \in [\epsilon^*,1]
\end{equation}

where the variables $\theta_s, \theta_l, \tau, \epsilon, \epsilon^*$ correspond to the the dimensionless forms of the temperature of solid phase, the temperature of liquid phase, time, spatial coordinate and the instantaneous interface position respectively.

We assume that the composition in the solid ($C_s$) is uniform and is equal to $k_0 C_l$, where  $k_0$ is the partition coefficient and $C_l$ is the liquid composition $C_s$. The governing PDE for $C_l$ can be written as follows:
\begin{equation}
\quad \frac{\partial C_l}{\partial\tau} = \frac{1}{Le}\frac{\partial^2C_l}{\partial\epsilon^2}\quad for\quad \epsilon \in [\epsilon^*,1]
\end{equation}
where $Le$ denotes the Lewis number which represents the ratio of thermal diffusivity and species mass diffusion coefficient.

The PDE for the interface heat balance (also known as Stefan's condition) is given by: 
\begin{equation}
    \quad \frac{1}{Ste}\frac{\partial\epsilon^*}{\partial\tau} = \frac{\partial\theta_s}{\partial\epsilon}-\frac{\partial\theta_l}{\partial\epsilon}\quad at\quad \epsilon = \epsilon^*
\end{equation}

Here, $Ste$ denotes the Stefan number which is a constant computed as per the formula:
\begin{equation}
    Ste = c_p(T_\infty-T_0)/L_f
\end{equation}
where $c_p$ is specific heat, $L_f$ is latent heat, $T_\infty$ is initial temperature of liquid and $T_0$ is temperature of solid at the origin.

The physics of diffusion gives rise to a boundary condition at the interface given by the PDE:
\begin{equation}
    \quad\frac{-1}{Le}\frac{\partial C_l}{\partial\epsilon}=C_l^*(1-k_0)\frac{d\epsilon^*}{d\tau}\quad at\quad \epsilon = \epsilon^*
\end{equation}

Here $k_0$ is the partition coefficient which is a constant computed as:
\begin{equation}
    k_0=C_s^*/C_l^*
\end{equation}
where $C_s^*$ is the interface solid composition and $C_l^\ast$ is the interface liquid composition.

The initial conditions are expressed as:
\begin{equation}
    \theta_l=1\quad at \quad \tau=0
\end{equation}
\begin{equation}
    \epsilon^*=0\quad at \quad \tau=0
\end{equation}
\begin{equation}
    C_l=C_0\quad at \quad \tau=0
\end{equation}
where $C_0$ is the composition of the alloy which is entirely liquid in its initial state implying the solid-liquid interface is initially at the origin.

There are boundary conditions at the origin and interface given by:
\begin{equation}
    \theta_s=0\quad at\quad \epsilon=0
\end{equation}
\begin{equation}\label{eq:12}
    \theta_s=\theta_l=\theta^*\quad at\quad \epsilon = \epsilon^*
\end{equation}
\hfil indicating that at the interface, the temperature of the two phases will always be equal\hfil

The phase diagram (Fig \ref{fig:fig1}(b)) couples the temperature field to the composition field. Assuming a linear form for the liquidus curve (for simplicity) as $T = T_f + m_lC_l$ allows expressing the interface liquid composition in terms of the interface temperature by:
\begin{equation}
    C_l^* = (T_0-T_\infty)(\theta_f-\theta^*)/m_l
\end{equation}
where $m_l$ is the liquidus slope and $\theta_f$ is the scaled dimensionless variable corresponding to the equilibrium freezing temperature $T_f$.

\paragraph{}Thus, now the problem consists of predicting the distribution of five unknowns in space ($\epsilon$) and time ($\tau$): $\theta_s(\epsilon,\tau), \theta_l(\epsilon,\tau), C_s, C_l(\epsilon,\tau), \epsilon^*(\tau)$ and requires solving five PDEs along with satisfying the initial and boundary conditions and maintaining other known relationships. In addition, an interface discontinuity is expected in the composition profile as indicated in Fig \ref{fig:fig1}(a). This sets up a challenging problem which can be solved with an approach utilising the PINN framework as described in the next section.

\section{Solution Methodology}\label{sec:solution}
\subsection{Conventional PINN Approach}
\paragraph{}Here, we discuss the conventional PINN approach for solving the solidification problem presented in the previous section. There is a set $ X = (\epsilon, \tau) $ corresponding to the independent variables: location and time and a set of dependent variables $ Y = (\theta, C, \epsilon^*) $ corresponding to the temperature, composition and interface position. The solution $Y$ for a given $X$ can be predicted using a single feed-forward multi-layer neural network $N$. There will be two neurons in the input layer for the set $X$ and $h$ neurons in each of the $n$ hidden layers. In order to predict for each phase separately, it is helpful to have five neurons in the output layer corresponding to the set $Y2 = (\theta_s, \theta_l, C_s, C_l, \epsilon^*)$. The final outputs $\theta$ and $C$ will depend on where $\epsilon$ lies with respect to $\epsilon^*$ and can be defined using Eqs. \ref{eq:14} and \ref{eq:15}.

\begin{equation}\label{eq:14}
\theta(\epsilon,\tau)=\theta_s(\epsilon,\tau)\mathds{1}_{(0,\epsilon^*(\tau))}(\epsilon)+\theta_l(\epsilon,\tau)\mathds{1}_{(\epsilon^*(\tau),1)}(\epsilon)
\end{equation}
\begin{equation}\label{eq:15}
C(\epsilon,\tau)=C_s(\epsilon,\tau)\mathds{1}_{(0,\epsilon^*(\tau))}(\epsilon)+C_l(\epsilon,\tau)\mathds{1}_{(\epsilon^*(\tau),1)}(\epsilon)
\end{equation}
\hfil where $\mathds{1}_{(0,\epsilon^*(\tau))}(\epsilon)$ denotes an indicator function taking a value of 1 if $\epsilon \in [0,\epsilon^*(\tau)]$ or 0 otherwise\hfil

\paragraph{}The loss function $L_{tot}$ which needs to be minimised in the PINN learning framework \cite{PINN} consists of various terms, listed as follows: (i) PDE residuals evaluated on collocation points sampled from the domain ($L_c$), (ii) the mean squared error (MSE) terms for labelled data used to enforce the initial conditions ($L_i$), (iii) boundary conditions ($L_b$) and (iv) mean squared error terms for a few measurement points ($L_m$) used to fit the available analytical solution. The requirement of a few measurement points arises due to the strong discontinuity in gradient of temperature and the concentration at the interface. Similar strategy was reported in other PINN approaches for moving boundary problems, \cite{Stefan} These terms are explicitly formulated below.

Let $p$ be the number of measurement data points. Then, the MSE losses for the 3 output variables will be:
\begin{gather}
L_{m1} = \frac{1}{p}\sum_{i=1}^p(\theta_i(\epsilon,\tau)-\theta_{anal}(\epsilon,\tau))^2\nonumber\\
L_{m2} = \frac{1}{p}\sum_{i=1}^p(C_i(\epsilon,\tau)-C_{anal}(\epsilon,\tau))^2\nonumber\\
L_{m3} = \frac{1}{p}\sum_{i=1}^p(\epsilon_i^*(\tau)-\epsilon_{anal}^*(\tau))^2\nonumber\\
L_m = L_{m1} + L_{m2} + L_{m3}\label{eq:16}
\end{gather}

Let $n_x$ and $n_t$ be the number of $\epsilon$ points and the number of $\tau$ points which make up the grid of collocation data. Then, there will be $n_c=n_x \times n_t$ number of collocation data points. Note that $p << n_c$.

The MSE losses for the prescribed 3 initial conditions will be:
\begin{gather}
    L_{i1} = \frac{1}{n_x}\sum_{i=1}^{n_x}(\theta_{l}(\epsilon_i,0)-1)^2\nonumber\\
L_{i2} = (\epsilon^*(0))^2\nonumber\\
L_{i3} = \frac{1}{n_x}\sum_{i=1}^{n_x}(C_l(\epsilon_i,0)-C_0)^2\nonumber\\
L_i = L_{i1} + L_{i2} + L_{i3}\label{eq:17}
\end{gather}

The MSE losses for the 4 constraints valid on evaluations at the boundary or interface will be:
\begin{gather}
    L_{b1} = \frac{1}{n_t}\sum_{i=1}^{n_t}(C_s(\epsilon^*,\tau_i)-k_0*C_l(\epsilon^*,\tau_i))^2\nonumber\\
L_{b2} = \frac{1}{n_t}\sum_{i=1}^{n_t}(\theta_s(0,\tau_i))^2\nonumber\\
L_{b3} = \frac{1}{n_t}\sum_{i=1}^{n_t}(\theta_s(\epsilon^*,\tau_i)-\theta_l(\epsilon^*,\tau_i))^2\nonumber\\
L_{b4} = \frac{1}{n_t}\sum_{i=1}^{n_t}(C_l(\epsilon^*,\tau_i) - (T_0-T_\infty)(\theta_f-\theta(\epsilon^*,\tau_i))/m_l)^2\nonumber\\
L_b = L_{b1} + L_{b2} + L_{b3} + L_{b4}\label{eq:18}
\end{gather}

The residual terms for the 5 PDEs in this problem (in order) are as follows. These terms involve the partial derivatives of the neural network outputs with respect to the inputs which have been computed using automatic differentiation capabilities of the tensorflow framework.
\begin{gather}
    L_{c1}=\frac{1}{n_c}\sum_{i=1}^{n_c} (\frac{\partial\theta_s^i}{\partial\tau} - \frac{\partial^2\theta_s^i}{\partial\epsilon^2})^2\nonumber\\
L_{c2}=\frac{1}{n_c}\sum_{i=1}^{n_c} (\frac{\partial\theta_l^i}{\partial\tau} - \frac{\partial^2\theta_l^i}{\partial\epsilon^2})^2\nonumber\\
L_{c3}=\frac{1}{n_c}\sum_{i=1}^{n_c} (\frac{\partial C_l^i}{\partial\tau} - \frac{1}{Le}\frac{\partial^2C_l^i}{\partial\epsilon^2})^2\nonumber\\
L_{c4}=\frac{1}{n_c}\sum_{i=1}^{n_c} (\frac{1}{Ste}\frac{\partial\epsilon_i^*}{\partial\tau} - \frac{\partial\theta_s^i}{\partial\epsilon}+\frac{\partial\theta_l^i}{\partial\epsilon})^2\nonumber\\
L_{c5}=\frac{1}{n_c}\sum_{i=1}^{n_c} (\frac{1}{Le}\frac{\partial C_l^i}{\partial\epsilon}+C_l^i(\epsilon^*,\tau)(1-k_0)\frac{d\epsilon_i^*}{d\tau})^2\nonumber\\
L_c=L_{c1}+L_{c2}+L_{c3}+L_{c4}+L_{c5}\label{eq:19}
\end{gather}

Thus, the total loss consists of a sum of 15 terms: 
\begin{equation}\label{eq:20}
    L_{tot}=L_m + L_i + L_b + L_c
\end{equation}

\paragraph{}The network $N$ is trained by minimising the total loss for the set of network parameters $W$ at the collocation points $X$ which are randomly sampled at each iteration. This can be done using standard optimisation techniques. However, training a single network, to model the behaviour of 5 different variables while maintaining 15 constraints, is difficult. Past studies have failed to reduce the loss to acceptable levels while solving problems of even lower complexity using the standard PINN approach. Hence, more recent methodologies were attempted to address the concerns. The formulation of these methods is provided in the following section with a sole purpose of assisting the reader in understanding the variations that led to the proposed model.


\subsection{Advances in PINN approaches}
\paragraph{}For a fully-connected feed-forward network, the number of hidden layers $h$ and the number of neurons in each layer $n$ are hyperparameters which ideally need to be optimised to solve a specific problem. The technique of bayesian optimisation was used for hyperparameter tuning of PINNs for solving Helmholtz problems in \cite{tuning}. However, in case of more complex problems, the optimisation of network parameters to sufficiently minimise the loss itself is difficult. Thus, here, the additional hyperparameter optimisation is eliminated for simplicity and $h=5$ and $n=100$ is taken which is same as in the original PINN work \cite{PINN}. The parameters of the network are initialised by the Glorot Normal method \cite{glorot} and updated using the Adam's optimisation scheme \cite{adam}. The Adam Optimisation combines the adaptive learning rate and momentum methods which results in accelerated convergence and has been popularly used for minimising the PINN loss as in \cite{SciML}, \cite{Burgers}, \cite{Biot}, \cite{Stokes}, \cite{Stefan}, \cite{Curing}. The activation function employed after each hidden layer is chosen to be Swish \cite{swish} (Eq. \ref{eq:21}). This function was chosen since its usage in PINN has resulted in superior convergence and higher accuracies as recorded while solving the Helmholtz Equation \cite{pinn_swish} and fluid flow problems \cite{resnet}.

\begin{equation}\label{eq:21}
    Swish(x)=xSigmoid(x)
\end{equation}

\paragraph{}A two-phase Stefan problem involving dynamic interactions during heat transfer between a solid and a liquid phase of a pure system separated by a free interface was solved using PINN in \cite{Stefan}. The system with missing conditions consisted of 3 PDEs: one heat equation for each phase and one for interface physics. They employ two neural networks: one to track the position of the interface between the phases and another to model the two temperature distributions. All constraints are added as error terms to the loss function and the two networks are simultaneously trained until the predicted evolution of temperature distribution and interface position sufficiently follows the analytical solutions. Taking cue from here, we train 3 networks to solve the solidification problem (each model specially catering to a physical variable). However, the temperature and composition predictions at a particular time turn out completely inaccurate, with a negative value being predicted for interface position. The predicted solution gets stuck in this intermediate state as the loss no longer changes after decreasing for initial iterations. Conventional PINN models are known to struggle in such transient problems as the network gets trapped in an incorrect local minimum which makes further optimisation difficult. The authors in \cite{Causality} reported the same issue while solving the 1D Allen-Cahn Equation. After careful diagnosis, they concluded that this happens because the loss is minimised even when the predictions at previous time steps are inaccurate thereby violating temporal causality. Thus, the causal training approach was introduced to ensure that training for future time steps proceeds only after the model has accurately learnt the behaviour at earlier times. It is described next in context of the solidification problem.

\paragraph{}Consider a discrete data by taking a uniform mesh of size $n_t$ x $n_x$ in the domain of $\tau$ and $\epsilon$ forming the set of collocation points which are randomly sampled at every iteration of training. If the loss at every time step $\tau_i$ averaged over all $\epsilon$ points as a function of network parameters $W$ is given by $L(\tau_i,W)$ then Eq. \ref{eq:22} defines the weighted time-average loss as per the causal training algorithm \cite{Causality}. A temporal weight $w_i$ is introduced for the loss term at every time step $\tau_i$ which is expected to be such that it is very large – allowing minimisation of $L(\tau_i,W)$ – only if all losses before $\tau_i$ are minimised to the best extent. This is achieved when $w_i$ takes the form given by Eq. \ref{eq:23} which implies that only if the sum of losses at previous times is small enough, the inverse exponential can make the weights large enough. Thus, this algorithm shall be effective in learning the temporal evolution of temperature distribution, composition distribution and interface position in the solidification problem. Note that the $e$ in the expression of $w_i$ is a causality parameter which controls the slope of $w_i$ and can be gradually increased after every $S$ iterations of training to increase the strength with which the PDE residual constraint is enforced across the temporal domain.

\begin{equation}\label{eq:22}
    L(W)=\frac{1}{n_t}\sum_{i=0}^{n_t}w_iL(\tau_i,W)
\end{equation}
\begin{equation}\label{eq:23}
    w_i=exp(-e\sum_{k=1}^{i-1}L(\tau_i,W))
\end{equation}

\paragraph{}The total loss ($L_{tot}$) in Eq. \ref{eq:20} could also be calculated as a weighted sum of its 15 component loss terms. The weights for these loss terms could be taken as hyperparameters whose values will be determined from training or they could be assigned values inversely proportional to the magnitude of contribution made by the corresponding loss term to the total. Fixed scaling is helpful to balance the loss terms apriori while adaptively varying weights provide the flexibility of auto-balancing at the cost of additional parameter optimisation. Loss terms which have been calculated with fewer points could be assigned higher weights so that they are enforced equally-well as compared to other terms. This was employed in \cite{Causality} to balance the initial condition loss and the residual loss. Along similar lines, the causal training approach with scaled loss terms was implemented to solve the solidification problem and the results after 1000 epochs of training with a learning rate of 0.01 are shown in Fig \ref{fig:fig8} (see Appendix). The network for $\epsilon^*$ was able to predict the direction and nature of interface movement but the predicted values were still offset from the analytical values. At a particular instance of time, the predicted temperature profile closely aligned with the analytical solution apart from a small discontinuity near the interface but the predicted composition profile was still inaccurate. This prompts a larger focus on modelling the composition especially due to its low values and the expected non-linear profile in the liquid phase having a high jump (discontinuity) at interface.


\paragraph{}An approach for adaptive weighting of the loss terms is the learning rate annealing put-forth in \cite{adaptive} which balances the interplay between the residual loss term and the different data-fit loss terms. The authors identified that a common mode of failure in PINNs is the one related to unbalanced gradients during backpropagation. To create a remedy to overcome this pathology, they took inspiration from the popular Adam’s optimiser in making use of the backpropagated gradient statistics computed during model training to automatically tune the weights. Consider the loss function for a parameter set $\theta$ defined in Eq. \ref{eq:24} as the sum of the residual loss and the weighted combination of M data-fit terms. First, the instantaneous weights ($\hat{\lambda_i}$) are calculated by computing the ratio between the maximum gradient of the residual loss and the mean of the gradient magnitudes of the data-fit loss under consideration over $\theta$ as given in Eq. \ref{eq:25}. Then, the actual weights $\lambda_i$ are computed as the running average of previous values (Eq. \ref{eq:26}) to reduce stochasticity arising from each gradient descent update. Finally, the gradient descent update takes place to update network parameters $\theta$ using weight values $\lambda_i$ (Eq. \ref{eq:27}) with a learning rate of $\eta$.


\begin{equation}\label{eq:24}
    L(\theta) = L_r(\theta) + \sum_{i=1}^M\lambda_i L_i(\theta)
\end{equation}
\begin{equation}\label{eq:25}
    \hat{\lambda_i} = \frac{max_{\theta}(|\nabla_{\theta} L_r(\theta)|)}{mean_{\theta}(|\nabla_{\theta} L_i(\theta)|)}
\end{equation}
\begin{equation}\label{eq:26}
    \lambda_{i,new} = (1-\alpha)*\lambda_{i,old}+\alpha*\hat{\lambda_i}
\end{equation}
\begin{equation}\label{eq:27}
    \theta_{new} = \theta_{old}-\eta\nabla_{\theta}L_r(\theta)-\eta\sum_{i=1}^M\lambda_{i,new}\nabla_{\theta} L_i(\theta)
\end{equation}

\paragraph{} Authors in \cite{Curing} use the PINN framework to simulate the thermochemical curing process of a composite material on a tool in an autoclave. The physics of this problem is expressed by a set of coupled non-linear PDEs describing heat conduction and resin cure kinetics. Loss minimisation became difficult when a single PINN is used to capture both the processes, hence, separate networks were trained: one to model the degree of cure ($\alpha$) and another to predict the temperature ($T$). However, when the simultaneous training of the 2 networks fails to converge they adopt a sequential training strategy. They minimise the $\alpha$-related losses for 30 epochs followed by the $T$-related losses for 30 epochs, repeating this procedure for 10 iterations until both the losses are reduced to acceptable levels. Authors in \cite{Stokes} propose a parallel-network architecture for solving the coupled Stokes-Darcy system where each network caters to a different sub-region of the solution. This would be helpful especially in problems having discontinuity at the interface when the solutions of the two regions on either side of the interface are very different. Indeed, using two networks resulted in lower error in predictions.

\subsection{Methodology Implemented}
\begin{figure}
  \centering
  \includegraphics[scale=0.4]{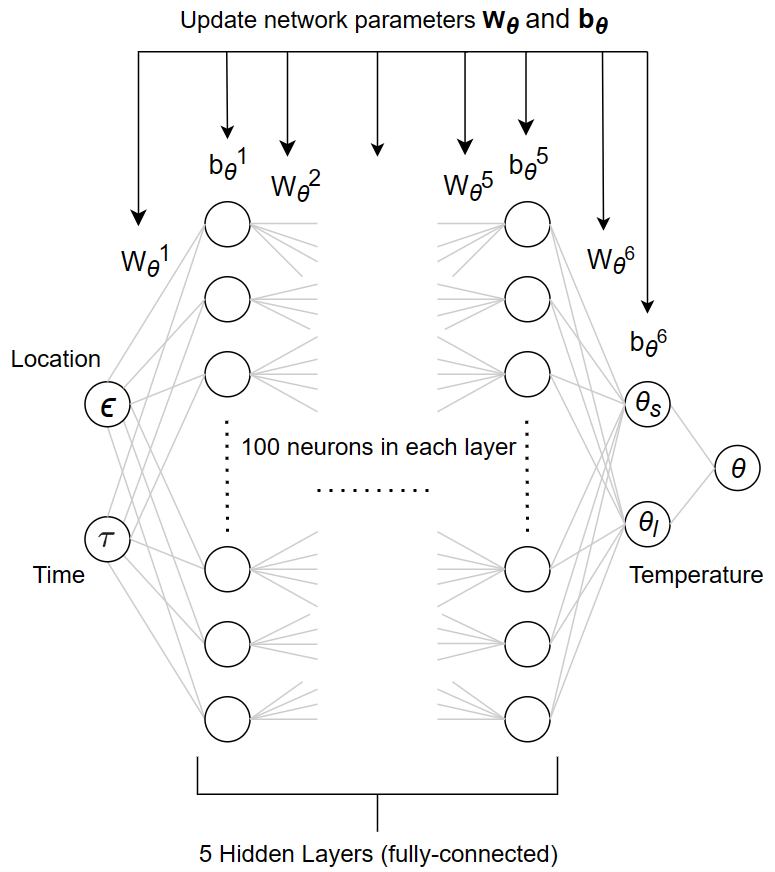}
  \includegraphics[scale=0.4]{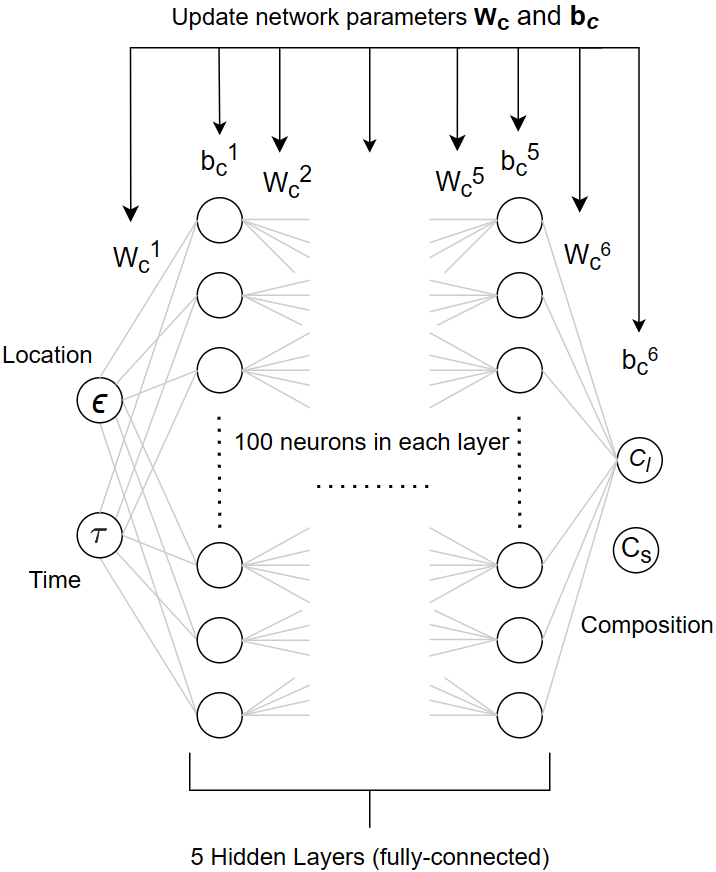}
  \includegraphics[scale=0.4]{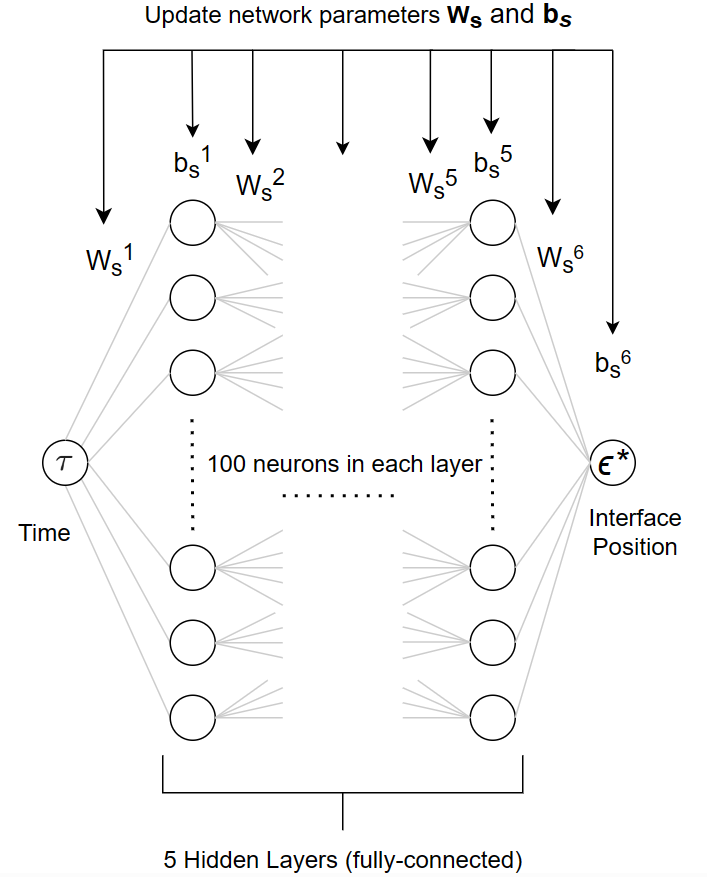}
  \caption{Schematic representations of the architectures of the 3 neural networks for (a) temperature distribution in both phases, (b) liquid concentration, and (c) inteface position}
  \label{fig:fig2}
\end{figure}

\paragraph{}A separate network can be considered for each physical variable and each phase based on the works of \cite{Stokes}, \cite{Stefan}, \cite{Curing}. However, taking 5 networks for predicting the 5 variables: $\theta_s(\epsilon,\tau), \theta_l(\epsilon,\tau), C_s, C_l(\epsilon,\tau), \epsilon^*(\tau)$ would lead to a computationally expensive training procedure. The problem is simplified using the information that $C_s$ is a constant and is therefore taken as a separate trainable parameter instead of using a neural network to approximate it. The task of capturing the interface jump discontinuity and modelling the nonlinear composition profile in the liquid phase is left to a neural network catered to predict $C_l(\epsilon,\tau)$. A single network had sufficiently modelled the linear temperature profiles during the implementation of causal training, therefore, a separate network for predicting the temperature of each phase is not needed. There will be a network required to track the interface movement. Thus, only 3 neural networks in total are taken with each predicting one of $\theta(\epsilon,\tau), C_l(\epsilon,\tau)$ and $\epsilon^*(\tau)$. Each network consists of $h=5$ hidden layers each having $n=100$ neurons and swish activation. The architectures are represented in Fig \ref{fig:fig2}.

\paragraph{}Let $W_s$, $W_\theta$ and $W_C$ denote the parameters of the networks for predicting interface position, temperatures and liquid composition respectively. Thus, in this problem there are 4 sets of variables to be optimised: the parameter $C_s$ and the parameter sets $W_s, W_\theta, W_C$ for the 3 networks. If all are optimised simultaneously, due to the inherent stochasticity of the training process and the complexity of the problem, the convergence of all the variables to the exact solution is difficult. It is easier to get an estimate of $C_s$ as an average over just a few composition measurements from anywhere in the solid phase without the need to enforce any physical constraints on it. Such an estimate (here the analytical value is taken) can be used to stop training for $C_s$ early and focus on optimising the 3 networks.

\paragraph{}The domain considered for data sampling and evaluation is $\epsilon \in [0,1]$ and $\tau \in [0,10]$. 900 points from the domain, 50 points at the interface ($\epsilon=\epsilon^*_{anal}$) and 50 initial points ($\tau=0$) totalling $p=1000$ measurement points are randomly sampled prior to training. At every iteration of causal training, a grid of $n_x=256$ and $n_t=100$ totalling $n_c=25600$ collocation points is randomly sampled. Accordingly, the total loss with loss terms scaled in inverse proportion to the number of points used to compute them is given by the following equation:

\begin{equation}
L_{tot}=25.6L_m+25600L_{i2}+100(L_{i1}+L_{i3})+256L_b+L_c
\end{equation}

\paragraph{}For this problem, a variation of the adaptive weighting algorithm \cite{adaptive} is taken. Consider the total loss to be composed of separate data-fit terms for temperature and composition, and other terms as defined by the set of Eqs. \ref{eq:29}. The weights for the data-fit loss terms for the 2 variables $\theta$ and $C$ are varied based on the gradient statistics of their data-fit and residual components with respect to their network parameters (Eq. \ref{eq:30}) and running average (Eq. \ref{eq:26} with $\alpha=0.1$) as discussed earlier. The Adam's optimiser is applied for minimising $L_{tot}$ while using adaptive weighting for varying $\lambda_1$ and $\lambda_2$ to update the network parameters. At every training iteration, 10000 collocation points are randomly sampled from the domain. When adaptive weighting was employed alone, after 5000 iterations, it was found that the learning for temperature and composition was well-balanced but the models couldn't capture the temporal evolution especially of the interface position (see Fig \ref{fig:fig9} in Appendix). This necessitates the use of causal training to improve temporal learning. However, causal training alone had failed to capture the complex composition profiles. Hence, using it along with adaptive weighting can be helpful to auto-focus more on the composition modelling.

\begin{gather}
L_{tot}=\lambda_1L_{ud}+\lambda_2L_{cd}+L_{others}\nonumber\\
L_{ud}=25.6L_{m1}\quad and\quad L_{cd}=25.6L_{m2}\nonumber\\
L_{others}=25.6L_{m3}+25600L_{i2}+100(L_{i1}+L_{i3})+256L_b+L_c\label{eq:29}
\end{gather}
\begin{gather}
    L_{ur}=L_{c1}+L_{c2}\quad and\quad L_{cr}=256L_{b1}+L_{c3}\nonumber\\
\hat{\lambda_1} = \frac{max_{W_\theta}(|\nabla_{W_\theta} L_{ur}(W_\theta)|)}{mean_{W_\theta}(|\nabla_{W_\theta} L_{ud}(W_\theta)|)}\quad and\quad \hat{\lambda_2} = \frac{max_{W_C}(|\nabla_{W_C} L_{cr}(W_C)|)}{mean_{W_C}(|\nabla_{W_C} L_{cd}(W_C)|)}\label{eq:30}
\end{gather}
\vspace{1pt}

\paragraph{}The sequential training strategy adopted by \cite{Curing} looked a promising option to employ causal training and adaptive weighting alternately and repeatedly so that both temporal learning and composition modelling are improved upon together. From the earlier discussions on causal training, with the dependency of $\theta$ and $C$ on $\epsilon^*$ evident from Eqs. \ref{eq:14} and \ref{eq:15}, it is clear that it is important for the PINN framework to first learn the interface movement fairly good enough for it to be able to predict the temperature and composition profiles accurately. Thus, after few iterations of causal training, the interface network can be frozen with weights at the epoch with the lowest error. Thereafter, Adam's optimisation with adaptively weighted loss function will optimise only the temperature and composition networks.

\begin{figure}[H]
  \centering
  \includegraphics[scale=0.8]{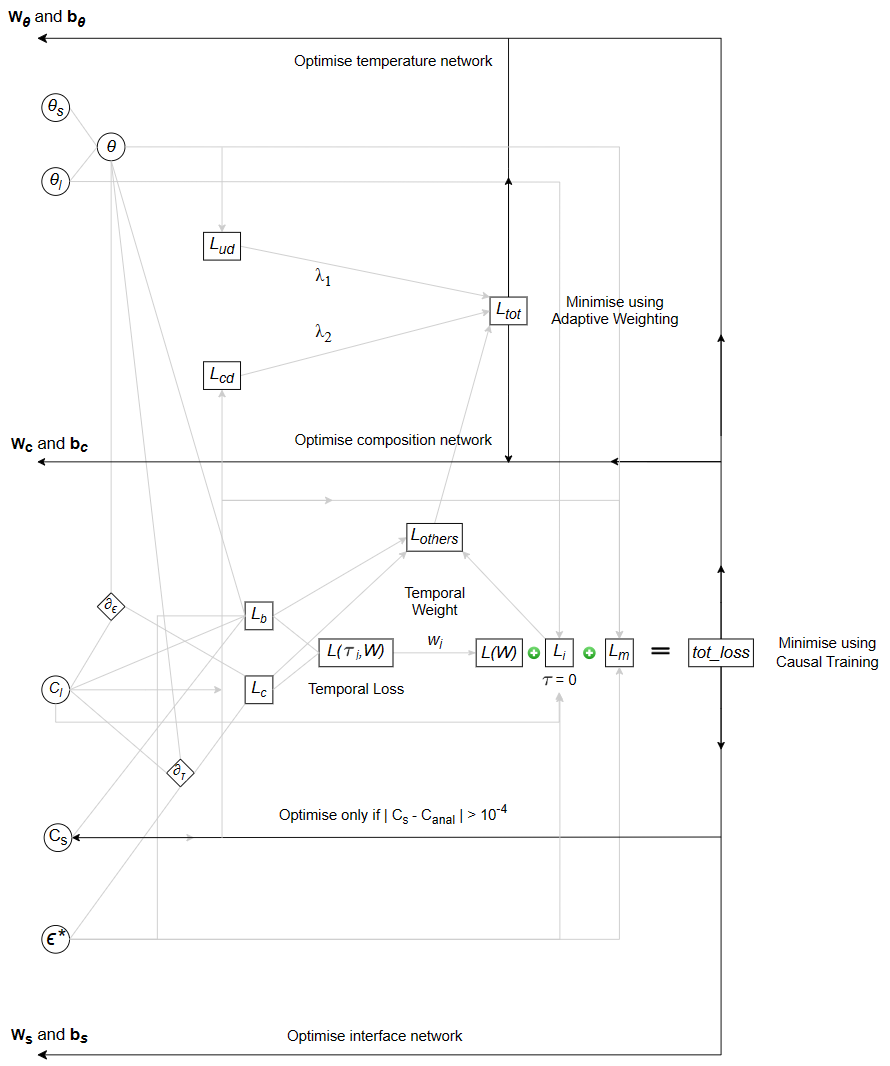}
  \caption{Schematic representation of the physics-informed component of the architecture for solving the solidification problem showing the make up of loss terms and the optimisation loops}
  \label{fig:fig3}
\end{figure}

\paragraph{}Finally, after studying various number of iterations and learning rate combinations, the following final scheme was adopted. Causal training was employed with Adam's optimisation for a total of 1000 epochs with a constant learning rate of $5\times 10^{-3}$. The loss function (Eq. \ref{eq:22}) was minimized with the causal parameter taking values from the set $[0.01,0.1,1,10,100]$ each for 200 epochs. At every epoch, Eq. \ref{eq:20} was taken as a representation of error and its value was noted. An error bound was kept as a stopping condition on training of $C_s$. Whenever, the absolute difference between the predicted and analytical value of $C_s$ went below $10^{-4}$, the parameter $C_s$ was fixed and training continued for the 3 networks. After 1000 epochs, the epoch with the lowest error was noted and the 3 networks were assigned weights from this epoch. Hereafter, the interface network was fixed while the other 2 networks were trained using Adam's optimisation with adaptively weighted loss function ($L_{tot}$ from Eq. \ref{eq:29}) for 1000 iterations with a constant learning rate of $10^{-4}$. Again, the epoch with the lowest error was noted and the 2 networks were assigned weights from this epoch. Now, another round of causal training is employed, this time with a lower learning rate of $5\times 10^{-4}$ for just 150 iterations with fixed $e=0.01$. The 2 networks take weights from the epoch with the lowest error and again undergo optimisation with adaptive weighting for 1000 epochs with same learning rate as earlier.

\paragraph{}The final scheme is outlined below which can be generalised to solve in an accelerated manner complex problems with coupled PDEs, multiple constraints, multiple governing parameters, multiple regions of solutions with discontinuities. This includes the pseudocodes for the specific application of causal training and adpative weighting algorithms considered here. Fig \ref{fig:fig3} pictorially shows the methodology depicting how the parameters of the 3 neural networks are optimised using feedback loops from the loss functions which enforce the physics and other conditions and are computed from outputs of these networks. The overall sequence of computations is given below. 

\begin{enumerate}
\item Design 3 networks to model $\theta(x,t), C_l(x,t), \epsilon^*(t)$
\item Take a trainable parameter $C_s$ with initial value 0.2
\item Sample $p=1000$ labelled data points prior to training
\item Perform causal training iteration: \cref{alg:ct}
\item Freeze the interface network $\epsilon^*(\tau)$
\item Perform adaptive weighting iteration: \cref{alg:aw}
\item Repeat \cref{alg:ct} and \cref{alg:aw} alternatively till acceptable convergence
\end{enumerate}

\begin{algorithm}[H]
    \caption{The Causal Training Iteration}
    \label{alg:ct}
\begin{algorithmic}
\FOR{ e = 0.01, 0.1, 1, 10, 100 }
    \FOR{ epoch = 1, 2, 3 ... S (S = 200) }
        \STATE Sample a grid of $n_x$ = 256 x $n_t$ = 100 points
        \STATE Compute $L_m$ (Eq. \ref{eq:16}) and $L_i$ (Eq. \ref{eq:17})
        \STATE Initialise $tot\_loss \gets 25.6L_m + 25600L_{i2} + 100(L_{i1}+L_{i3})$
        \FOR{ i = 1, 2, 3 ... $n_t$}
            \STATE Compute $L_b$ (Eq. \ref{eq:18}) and $L_c$ (Eq. \ref{eq:19})
            \STATE Compute temporal loss $L(\tau_i,W) \gets 256L_b + L_c$
            \STATE Compute temporal weight $w_i$ (Eq. \ref{eq:23})
        \ENDFOR
        \STATE Compute weighted time-average loss $L(W)$ (Eq. \ref{eq:22})
        \STATE Update $tot\_loss \gets tot\_loss + L(W)$        
        \STATE Compute gradients of loss w.r.t network parameters
        \IF{$|C_s(pred)-C_s(anal)|\le10^{-4}$}
            \STATE Stop training for $C_s$
        \ENDIF
        \STATE Use Adam's optimisation to minimise $tot\_loss$ and update parameters
    \ENDFOR
\ENDFOR
\end{algorithmic}
\end{algorithm}

\begin{algorithm}
    \caption{The Adaptive Weighting Iteration}
    \label{alg:aw}
\begin{algorithmic}
        \STATE Initialise $\lambda_1 = \lambda_2 = 1000$
        \FOR{ epoch = 1, 2, 3 ... 1000 }
            \STATE Sample 10000 collocation points
            \STATE Compute $L_{tot}$ (Eq. \ref{eq:29})
            \STATE Compute gradients of loss w.r.t network parameters
            \STATE Compute instantaneous weights using gradient statistics (Eq. \ref{eq:30})
            \STATE Update $\lambda_1$ and $\lambda_2$ (Eq. \ref{eq:26} with $\alpha=0.1$) and recompute $L_{tot}$
            \STATE Use Adam's optimisation to minimise $L_{tot}$ and update parameters
        \ENDFOR
\end{algorithmic}
\end{algorithm}

\section{Results}\label{sec:results}

\begin{figure}[H]
  \centering
  \begin{subfigure}{0.48\textwidth}
      \includegraphics[width=\linewidth]{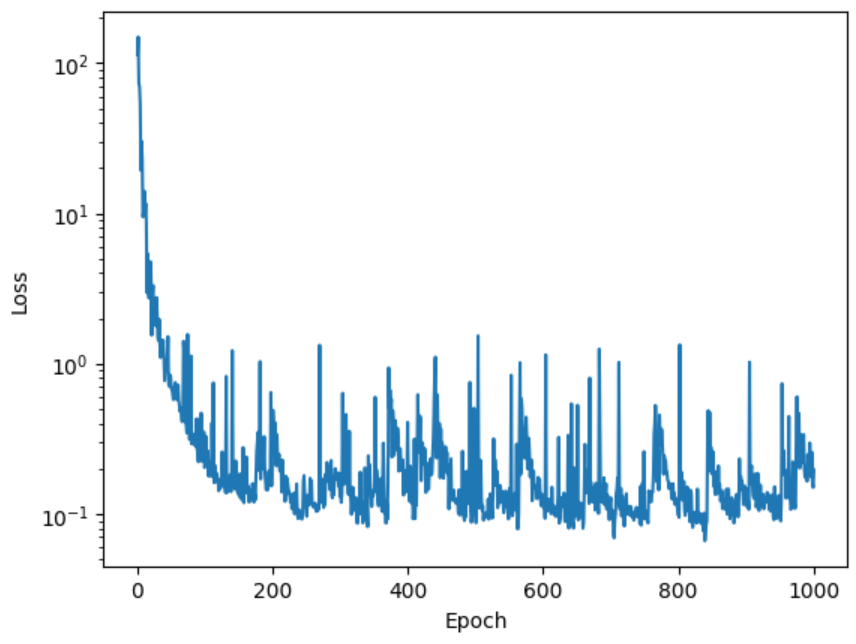}
      \caption{}
      \label{fig:subfig4a}
  \end{subfigure}
  \hspace*{\fill}
  \begin{subfigure}{0.48\textwidth}
      \includegraphics[width=\linewidth]{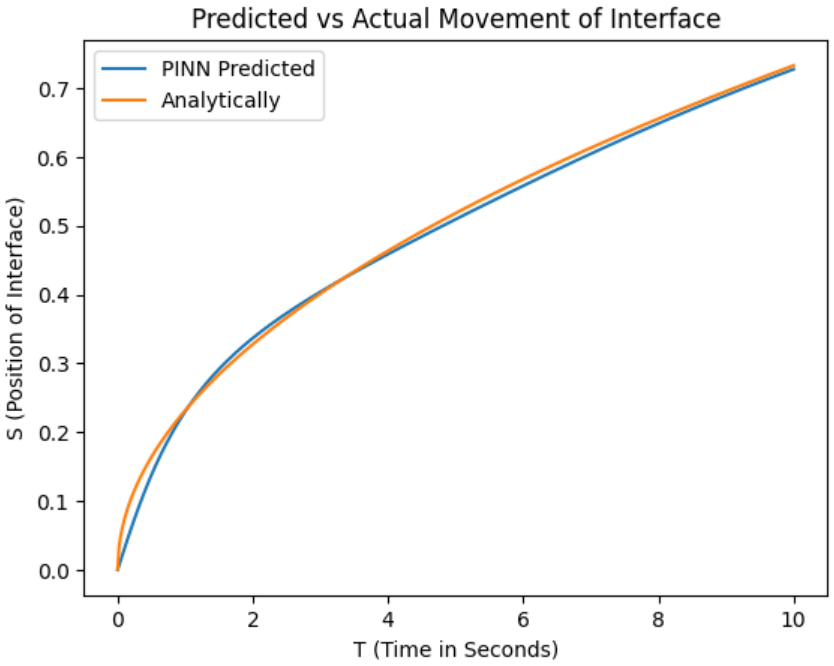}
      \caption{}
      \label{fig:subfig4b}
  \end{subfigure}
  \caption{Results of the proposed model showing (a) Loss curve, and (b)  interface movement after first iteration of causal training}
  \label{fig:fig4}
\end{figure}

\begin{figure}[H]
  \centering
  \begin{subfigure}{0.48\textwidth}
        \includegraphics[width=\linewidth]{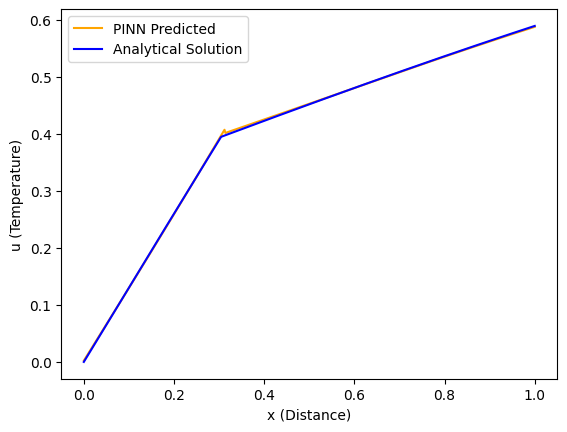}
        \caption{}
        \label{fig:subfig5a}
  \end{subfigure}
  \hspace*{\fill}
  \begin{subfigure}{0.48\textwidth}
        \includegraphics[width=\linewidth]{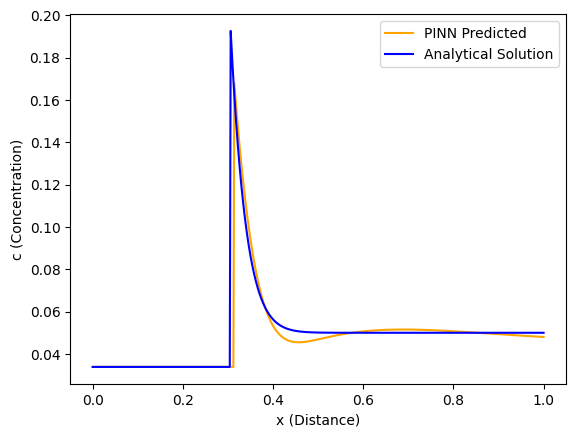}
        \caption{}
        \label{fig:subfig5b}
  \end{subfigure}
  \caption{(a) Predicted temperature distribution $\theta(x)$ showing a jump in the gradient at the interface, and (b) composition $C(x)$, compared with the analytical solution at  $t=1.725 $ s}
  \label{fig:fig5}
\end{figure}

\begin{figure}
  \centering
  \begin{subfigure}{0.49\textwidth}
        \includegraphics[width=\linewidth]{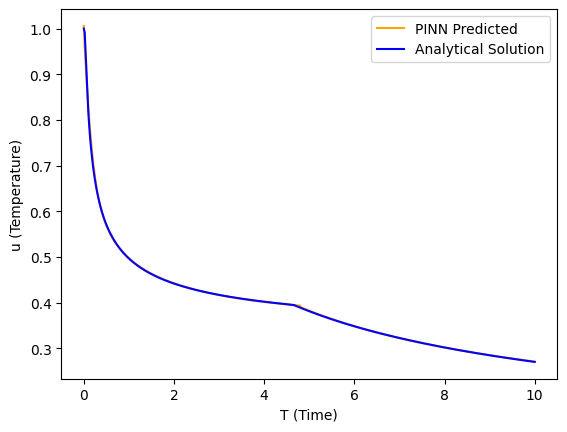}
        \caption{}
        \label{fig:subfig5c}
  \end{subfigure}
  \hspace*{\fill}
  \begin{subfigure}{0.49\textwidth}
        \includegraphics[width=\linewidth]{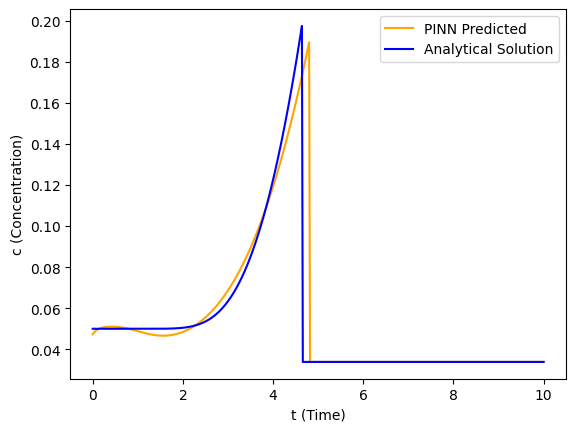}
        \caption{}
        \label{fig:subfig5d}
  \end{subfigure}
  \caption{Predicted transient evolution of the (a) temperature $\theta(t)$ and (b) composition $C(t)$, and the corresponding comparison with the analytical solution at $x=0.5$}
  \label{fig:fig10}
\end{figure}

\paragraph{}The final methodology described above is implemented as it is to solve the solidification problem. The training for $C_s$ stopped during the iterations of causal training, the absolute difference between the predicted and analytical value (0.03387) being $3.445 \times 10^{-5}$ indicating a desirable low error. The loss vs epoch curve for the 1000 iterations of causal training is shown in Fig \ref{fig:subfig4a}. Maximum amount of learning happened for $e=0.01$ (first 200 epochs) with marginal reduction in loss in further epochs. The lowest total error ($L_{tot}$ in Eq. \ref{eq:20}) was recorded at the 826$^{th}$ epoch. The predicted evolution of interface movement closely follows the analytical curve as seen in Fig \ref{fig:subfig4b}. The temperature and composition networks were then trained for 1000 epochs using adaptively weighted loss function. The lowest error/loss was recorded at the 688$^{th}$ epoch. From here, they were trained using causal training for 150 epochs with $e=0.01$ and the lowest error occured at the 76$^{th}$ epoch. From here, they were again trained for 1000 epochs using adaptively weighted loss function. Figs \ref{fig:fig5} and \ref{fig:fig10} show the final prediction results against the analytical solutions which closely align with each other (in particular space and time respectively). In contrast to employing different strategies on a standalone basis, the framework with this methodology is now able to capture the complex composition profile with its high peaks at interface along with a more accurate temperature profile and has improved in temporal learning across the 3 variables.

\begin{figure}
  \centering
  \begin{subfigure}{0.49\textwidth}
        \includegraphics[width=\linewidth]{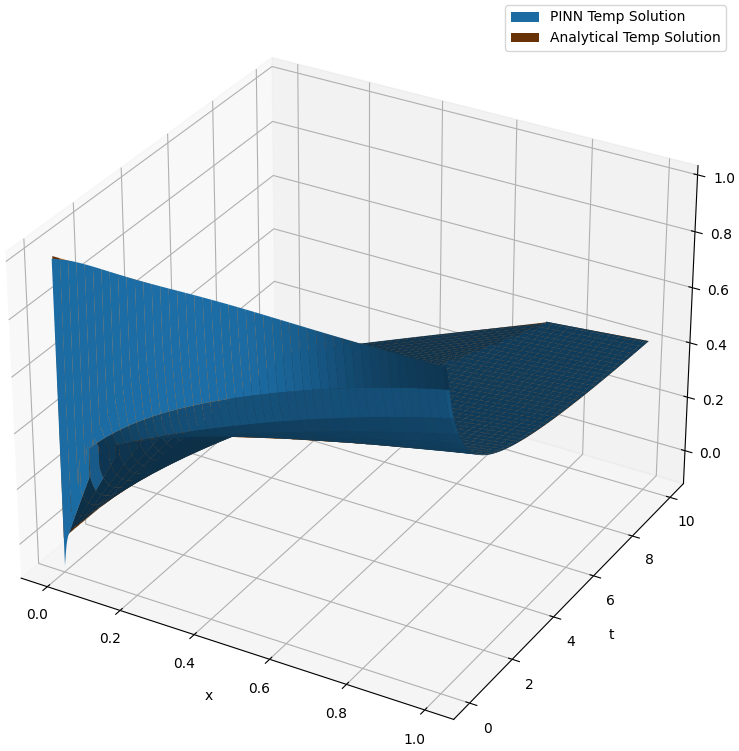}
        \caption{}
        \label{fig:subfig6a}
  \end{subfigure}
  \hspace*{\fill}
  \begin{subfigure}{0.49\textwidth}
        \includegraphics[width=\linewidth]{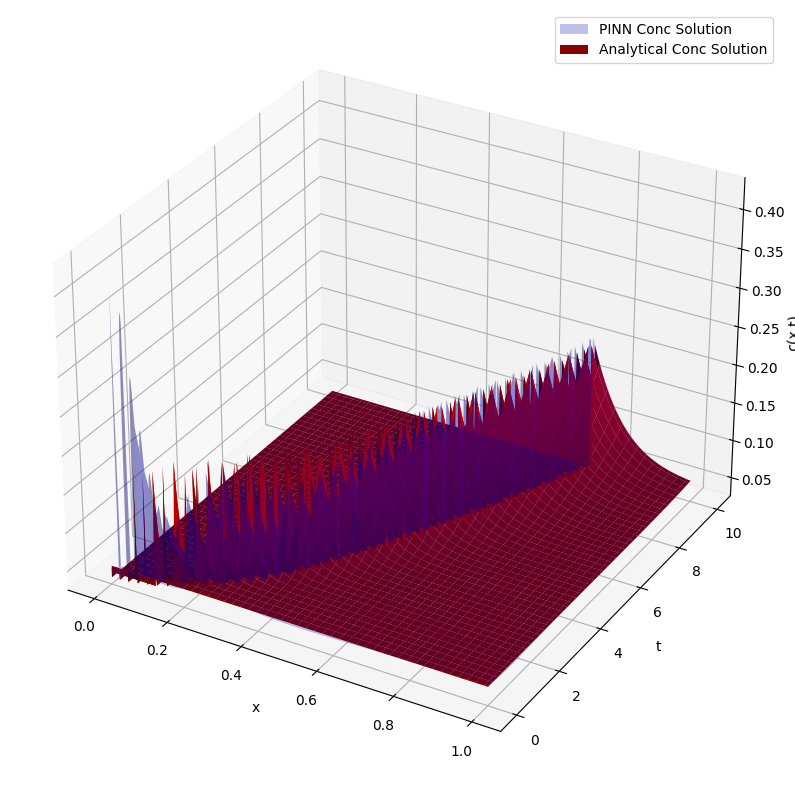}
        \caption{}
        \label{fig:subfig6b}
  \end{subfigure}  
  \caption{The predicted and analytical 3D plots for (a) temperature $\theta(x,t)$ and (b) composition $C(x,t)$. The horizontal plane shows the $(x,t)$ coordinate space  }
  \label{fig:fig6}
\end{figure}

\paragraph{}The temperature and composition predictions are plotted on a grid of 500x500 points along with the analytical solutions (Fig \ref{fig:fig6}). The temperature plots seem to completely overlap each other while the composition plots have an acceptable overlap although some predicted peaks are lower, and some are offset. The Mean Squared Errors (MSEs) and Relative $L^2$ Errors (RLE) between the predicted and analytical solutions for the 3 variables over the grid are noted in Eqs. \ref{eq:31}. They are quite low for temperature and interface position (given training for just 3150 epochs in total) while the $L^2$ error in composition is driven up by falsely predicted high peaks and the propagated error in interface position. Another run of the entire code resulted in lower error in composition due to a lower error in interface position as noted RLE2 in Eqs. \ref{eq:31}. Note that only 2 rounds of alternate application of causal training and adaptive weighting were employed here with limited number of epochs. It is expected that more rounds and iterations, and handling points nearby origin and interface, will improve the predictions further.

\begin{gather}
    MSE (\theta_{anal}, \theta_{PINN}) = 2.32\times10^{-5}, RLE (\theta_{anal}, \theta_{PINN}) = 1.24\%, RLE2 = 1.79\%\nonumber\\
MSE (C_{anal}, C_{PINN}) = 2.51\times10^{-4}, RLE (C_{anal}, C_{PINN}) = 28.11\%, RLE2 = 23.17\%\nonumber\\
MSE (\epsilon^*_{anal}, \epsilon^*_{PINN}) = 1.26\times10^{-4}, RLE (\epsilon^*_{anal}, \epsilon^*_{PINN}) = 2.16\%, RLE2 = 1.59\%\label{eq:31}
\end{gather}






\section{Conclusion}\label{sec:conc}
\paragraph{}In this work, we propose a PINN approach for solving a coupled, moving boundary system of PDEs. The model problem is that of a binary alloy solidification, involving heat and species transport, constrained by respective interface boundary conditions at the evolving interface. Firstly, conventional and the recent PINN approaches that were reported in the literature for uncoupled moving boundary problem were attempted. It was found that the approaches failed to capture the jump in the concentration field, which was an added complexity over the uncoupled moving boundary PDE. Based on these findings, we propose a novel combination of causal training and adaptive optimization, and its utility is demonstrated to model the coupled problem. An architecture consisting of separate independent networks was devised to cater to each output variable, namely temperature (in both phases), liquid composition and interface position, while a trainable parameter is taken for solid composition which is assumed to be a constant. The three networks and the parameter are trained simultaneously, while being constrained by the loss functions that embed the boundary and initial conditions along with the governing PDEs (physics) into the learning process. 

\paragraph{}The methodology adopted here implements the following four aspects useful for any problem in general: (i) treating each component of a multi-variable multi-phase problem separately (ii) temporal learning using causal training with scaled loss terms in a transient problem, alternatively applied with (iii) adaptive weighting of data-fit losses with variable separated terms while enforcing other constraints, and (iv) sequentially-focused training strategy to progressively narrow down the optimisation space. This results in good predictions (both in space and time) including a vastly-improved composition profile which is able to capture the non-linear profiles as well as the discontinuity at the interface (with full height at various times). A more accurate prediction of interface movement early lowers the error in composition as noted in the second run of the code. It is to be noted that here the causal training is the chief learning process which is time-consuming while the adaptive weighting can be considered as a quick supplementary fine-tuning process. The causal training and adaptative weighting could be employed using alternating iterative calculations to enable more accurate predictions. By virtue of remaining within the PINN framework, this approach can be employed when there are less labelled data points, experimentation is expensive, and computational cost of numerical solvers is high.

\section*{Acknowledgments}
The authors would like to acknowledge the financial support for this research provided via a scholarship by the Center for Machine Intelligence and Data Science at Indian Institute of Technology Bombay. 

\section*{Resources}
All relevant codes for this research are available in the form of .ipynb notebooks on github at \href{https://github.com/shiv12spingo/PINN_Research/tree/main/Solidification_Problem}{https://github.com/shiv12spingo/PINN\_Research/tree/main/Solidification\_Problem} and can be run on Google Colab. The implementation of the discussed methodology and the results can be found in Final\_Code.ipynb with the second run available in Code\_Run\_2.ipynb.

\newpage

\section*{Appendix}

\subsection{Analytical solution}
Following, \cite{book1} (section 5.2.2), the analytical solution of the binary alloy solidification of a superheated melt in a one-dimensional framework is given as follows.
\begin{itemize}
    \item Temperature profiles: \\
    \begin{gather}
    \theta_s = \frac{\theta^*}{\text{erf}(\phi)} \text{erf} \left( \frac{\phi \epsilon}{2 \sqrt{\tau}} \right)
    \\
    \theta_l = 1 - \frac{1-\theta^*}{\text{erfc}(\phi)} \text{erfc} \left( \frac{\phi \epsilon}{2 \sqrt{\tau}} \right)
    \end{gather}

    \item Interface position:\\
    \begin{equation}
        \epsilon^* = 2\phi \sqrt{\tau}
    \end{equation}

    \item Concentration of the liquid $C_{l}$:\\
    \begin{equation}
    C_{l} = C_{0}+\frac{C^{*}_{l}-C_{0}}{\text{erfc} (\phi \sqrt{Le})} \text{erfc} \left( \frac{\phi \sqrt{Le}}{2\sqrt{\tau}} \right)
    \end{equation}

    \item Temperature of the interface via the following transcendental equation: \\
    \begin{equation}
    \left[ \theta_f - \text{erf}(\phi) \left( 1+\frac{\sqrt{\pi}}{\text{Ste}} \phi \exp (\phi^2) \text{erfc}(\phi) \right) \right] \times \left[ 1- \sqrt{\pi Le}(1-k_0) \phi \exp (\phi^2 Le) \text{erfc}(\phi \sqrt{Le}) \right] = \frac{m_l C_0}{T_0-T_\infty}
    \end{equation}

\end{itemize}
In all the equations above, $\phi$ refers to the similarity parameter that scales interface position and square root of the time. 

\subsection{Predictions of the binary alloy solidification problem from existing PINN frameworks}
\begin{figure}[H]
  \centering
  \begin{subfigure}{0.33\textwidth}
        \includegraphics[width=\linewidth]{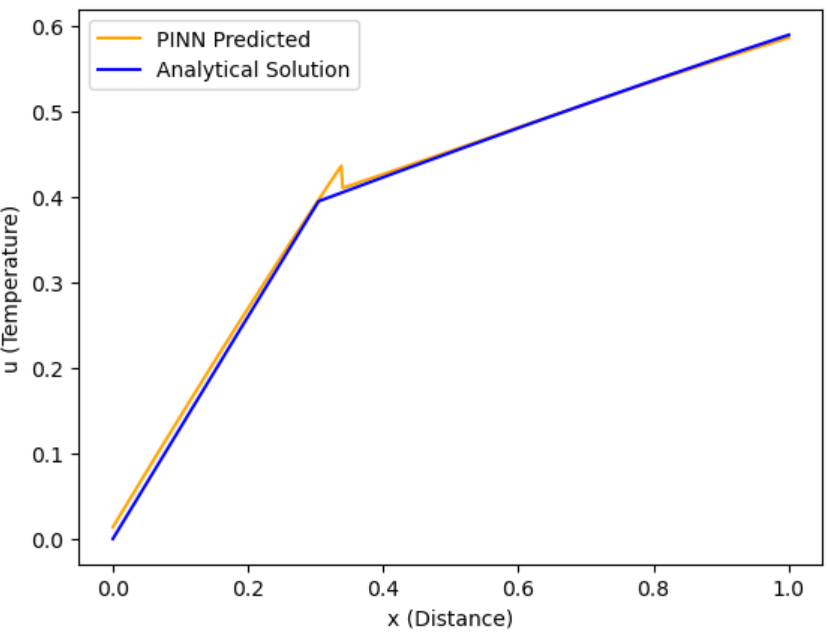}
        \caption{}
        \label{fig:subfig3a}
  \end{subfigure}
  \begin{subfigure}{0.33\textwidth}
        \includegraphics[width=\linewidth]{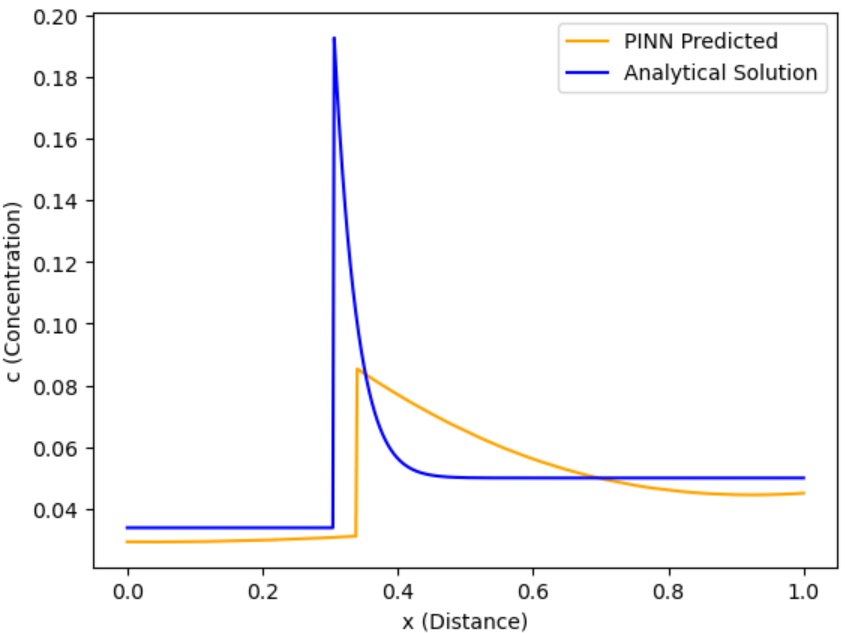}
        \caption{}
        \label{fig:subfig3b}
  \end{subfigure}
  \begin{subfigure}{0.33\textwidth}
        \includegraphics[width=\linewidth]{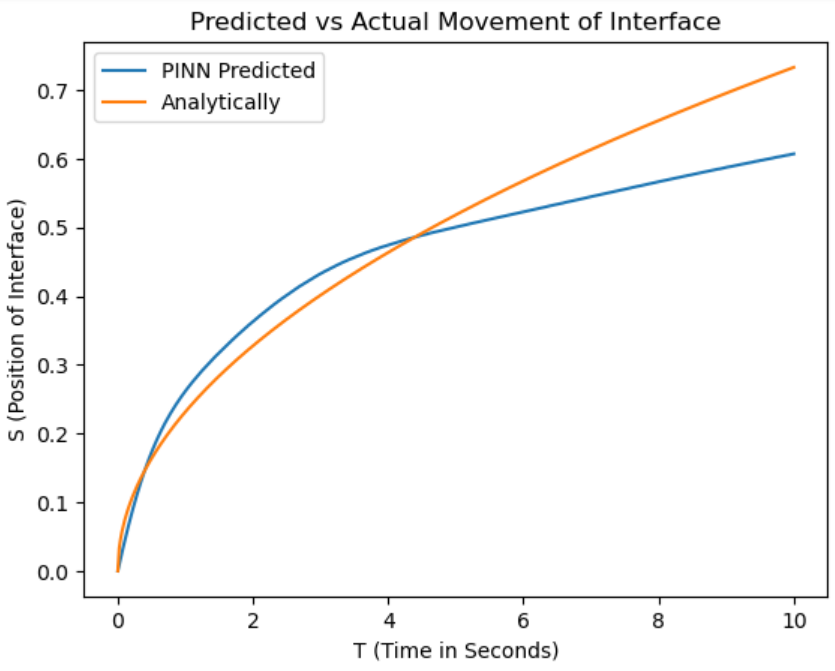}
        \caption{}
        \label{fig:subfig3c}
  \end{subfigure}
  \caption{Predicted temperature $\theta(x)$ (a) and composition $C(x)$ (b) profiles at $t=1.725s$ and the interface movement (c) after causal training are plotted against the analytical solution}
  \label{fig:fig8}
\end{figure}

\begin{figure}[H]
  \centering
  \begin{subfigure}{0.33\textwidth}
        \includegraphics[width=\linewidth]{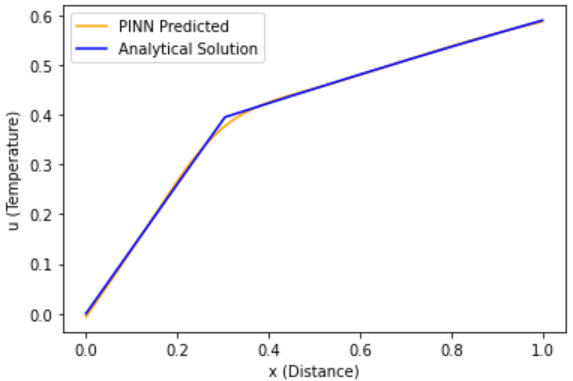}
        \caption{}
        \label{fig:subfig7a}
  \end{subfigure}
  \begin{subfigure}{0.33\textwidth}
        \includegraphics[width=\linewidth]{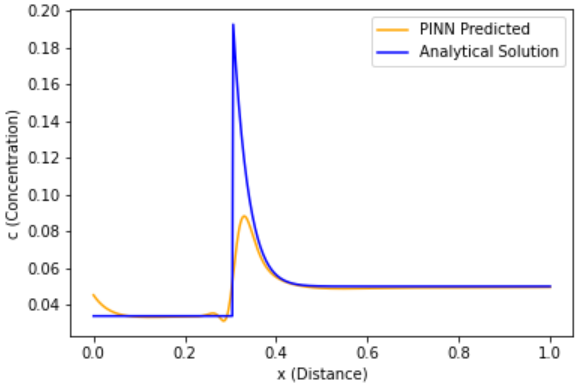}
        \caption{}
        \label{fig:subfig7b}
  \end{subfigure}
  \begin{subfigure}{0.33\textwidth}
        \includegraphics[width=\linewidth]{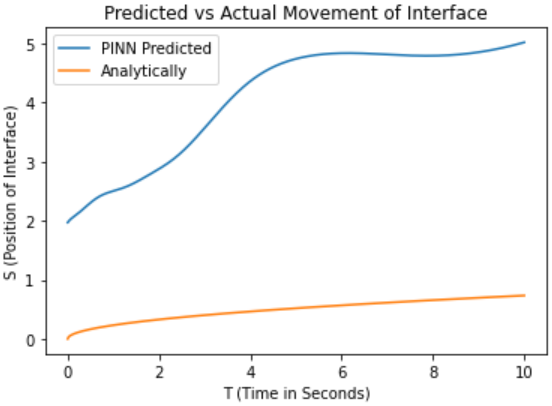}
        \caption{}
        \label{fig:subfig7c}
  \end{subfigure}
  \caption{Predicted temperature $\theta(x)$ (a) and composition $C(x)$ (b) profiles at $t=1.725s$ and the interface movement (c), after training with adaptively-weighted loss function (Eqs. \ref{eq:29} and \ref{eq:30}), are plotted against the analytical solution}
  \label{fig:fig9}
\end{figure}

The figures below show results from recent PINN based approaches that are obtained by causal training alone, see Figure \ref{fig:fig8} and adaptive weighting alone, see Figure \ref{fig:fig9}. It was found that when attempts were made to ensure close adherence to one of the variables in the coupled system, the other deviated significantly. This highlights the need for training the model through suitable combinations of the two approaches. 


\newpage
\bibliographystyle{unsrt}  
\bibliography{PINN_manuscript_SK_SK}

\end{document}